\documentclass[11pt,a4paper]{article}
\usepackage[hyperref]{emnlp2018}
\usepackage{times}
\usepackage{latexsym}
\usepackage{graphicx}
\usepackage{enumitem}

\usepackage{url}
\usepackage{multirow}
\usepackage{tikz}
\usepackage{amsmath}
\usepackage{todonotes}

\aclfinalcopy % Uncomment this line for the final submission
\title{Why Self-Attention?\\ A Targeted Evaluation of Neural Machine Translation Architectures}

\author{Gongbo Tang$^1$\thanks{\ \ Work carried out during a visit to the machine translation group at the University of Edinburgh.} \quad Mathias M\"{u}ller$^2$ \quad Annette Rios$^2$ \quad Rico Sennrich$^{2,3}$ \bigskip\\
  $^1$Department of Linguistics and Philology, Uppsala University\medskip\\
  $^2$Institute of Computational Linguistics, University of Zurich\medskip\\
  $^3$School of Informatics, University of Edinburgh}

\date{}

\begin{document}
\maketitle

\begin{abstract}
Recently, non-recurrent architectures (convolutional, self-attentional) have outperformed RNNs in neural machine translation. CNNs and self-attentional networks can connect distant words via shorter network paths than RNNs, and it has been speculated that this improves their ability to model long-range dependencies. However, this theoretical argument has not been tested empirically, nor have alternative explanations for their strong performance been explored in-depth. We hypothesize that the strong performance of CNNs and self-attentional networks could also be due to their ability to extract semantic features from the source text, and we evaluate RNNs, CNNs and self-attention networks on two tasks: subject-verb agreement (where capturing long-range dependencies is required) and word sense disambiguation (where semantic feature extraction is required). Our experimental results show that: 1) self-attentional networks and CNNs do not outperform RNNs in modeling subject-verb agreement over long distances; 2) self-attentional networks perform distinctly better than RNNs and CNNs on word sense disambiguation.  
\end{abstract}

\section{Introduction}
  \label{sec:introduction}

Different architectures have been shown to be effective for neural 
machine translation (NMT), ranging from recurrent architectures 
\cite{kal2013recurrent,bahdanau15joint,sutskever2014sequence,luong2015attention} 
to convolutional \cite{kal2013recurrent,gehring2017convolutional} 
and, most recently, fully self-attentional (Transformer) models 
\cite{vaswani2017Attention}.
Since comparisons \cite{gehring2017convolutional,vaswani2017Attention,Hieber2017sockeye} 
are mainly carried out via BLEU \cite{papineni2002}, it is inherently difficult to
attribute gains in BLEU to architectural properties.

Recurrent neural networks (RNNs) \cite{elman1990finding} can easily deal with
variable-length input sentences and thus are a natural choice for the encoder and
decoder of NMT systems. Modern variants of RNNs, such as GRUs \cite{cho2014learning}
and LSTMs \cite{hochreiter1997long}, address the difficulty of training recurrent
networks with long-range dependencies.
\newcite{gehring2017convolutional} introduce a neural architecture where both the
encoder and decoder are based on CNNs, and report better BLEU scores than RNN-based NMT models. 
Moreover, the computation over all tokens can be fully parallelized during training, which increases efficiency.
\newcite{vaswani2017Attention} propose Transformer models, 
which are built entirely with attention layers, without convolution or recurrence. 
They report new state-of-art BLEU scores for EN$\rightarrow$DE and EN$\rightarrow$FR. 
Yet, the BLEU metric is quite coarse-grained, and offers no insight as to which aspects
of translation are improved by different architectures.

To explain the observed improvements in BLEU, previous work has drawn on theoretical arguments.
Both \newcite{gehring2017convolutional} and \newcite{vaswani2017Attention} 
argue that the length of the paths in neural networks between co-dependent elements
affects the ability to learn these dependencies: 
the shorter the path, the easier the model learns such dependencies. 
The papers argue that Transformers and CNNs are better suited 
than RNNs to capture long-range dependencies.

However, this claim is based on a theoretical argument and has not been empirically tested.
We argue other abilities of non-recurrent networks could be responsible for their
strong performance. Specifically, we hypothesize that the improvements in BLEU are due to
CNNs and Transformers being strong semantic feature extractors.

In this paper, we evaluate all three popular NMT architectures: models based on RNNs (referred
to as \textit{RNNS2S} in the remainder of the paper), based on CNNs (referred to as \textit{ConvS2S}) and
self-attentional models (referred to as Transformers).
Motivated by the aforementioned theoretical claims regarding path length and semantic feature extraction,
we evaluate their performance on a subject-verb agreement task (that requires modeling long-range dependencies)
and a word sense disambiguation (WSD) task (that requires extracting semantic features). Both tasks build
on test sets of contrastive translation pairs, \textit{Lingeval97} \cite{sennrich2017grammatical}
and \textit{ContraWSD} \cite{rios2017improving}.

The main contributions of this paper can be summarized as follows: 
\begin{itemize}%[noitemsep]
  \item We test the theoretical claims that architectures with shorter paths 
    through networks are better at capturing long-range dependencies. 
    Our experimental results on modeling subject-verb agreement over 
    long distances do not show any evidence that Transformers or 
    CNNs are superior to RNNs in this regard.
  \item We empirically show that the number of attention heads in Transformers impacts their
    ability to capture long-distance dependencies. Specifically, many-headed multi-head attention is essential for modeling long-distance phenomena with only self-attention.
  \item We empirically show that Transformers excel at WSD, 
    indicating that they are strong semantic feature extractors. 
\end{itemize}

\section{Related work} 
  \label{sec:related_work}

\newcite{yin2017comparative} are the first to compare CNNs, LSTMs and GRUs 
on several NLP tasks. They find that CNNs are 
better at tasks related to semantics, while RNNs are better at 
syntax-related tasks, especially for longer sentences. 

Based on the work of \newcite{Linzen2016assessing}, 
\newcite{bernardy2017using} find that RNNs perform better than 
CNNs on a subject-verb agreement task, which is a good proxy 
for how well long-range dependencies are captured. 
\newcite{Tran2018recurrent} find that a Transformer language model performs worse
than an RNN language model on a subject-verb agreement task. They, too, note
that this is especially true as the distance between subject and
verb grows, even if RNNs resulted in a higher
perplexity on the validation set. This result of \newcite{Tran2018recurrent}
is clearly in contrast to the general finding that Transformers
are better than RNNs for NMT tasks. 

\newcite{bai2018empirical} evaluate CNNs and LSTMs on several 
sequence modeling tasks. They conclude that CNNs are better than 
RNNs for sequence modeling. However, their CNN models 
perform much worse than the state-of-art LSTM models on some 
sequence modeling tasks, as they themselves state in the appendix.

\newcite{Tang2018evaluation} evaluate different RNN architectures and Transformer models 
on the task of historical spelling normalization which translates a historical spelling into its modern form. 
They find that Transformer models surpass RNN models only in high-resource conditions.

In contrast to previous studies, we focus on the machine translation task, where architecture comparisons so far are mostly based on BLEU. 

%Most recently, \newcite{Domhan2018how} applies auxiliary techniques typically found in Transformers (multi-head attention, layer normalization, residual feed-forward layers) to NMT models with RNN or CNN architectures. If all of them are applied in the same fashion across architectures, they can no longer be a confounding factor for comparisons, and differences between models lie in the architectures themselves.

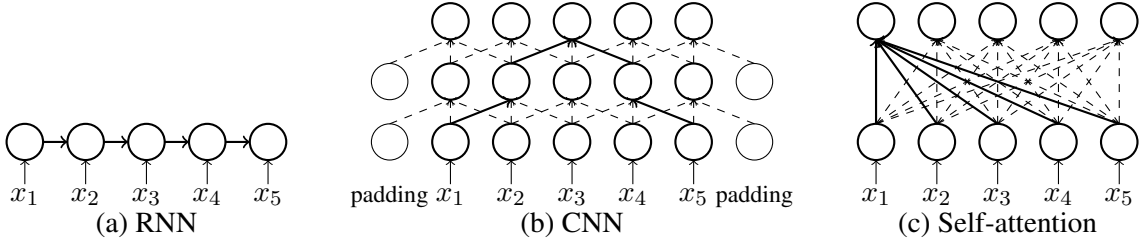
\begin{figure*}[htbp]
\centering
\begin{tikzpicture}[scale=0.8]
\draw [thick](1,1) circle [radius=0.3] (2,1) circle [radius=0.3] (3,1) circle [radius=0.3] (4,1) circle [radius=0.3] (5,1) circle [radius=0.3]; 
\draw [thick] [->] (1.3, 1) -- (1.7,1);
\draw [thick] [->] (2.3, 1) -- (2.7,1); 
\draw [thick] [->] (3.3, 1) -- (3.7,1);
\draw [thick] [->] (4.3, 1) -- (4.7,1); 
\draw [->] (1,0.3) -- (1,0.7); \node [below] at (1,0.4) {$x_{1}$};
\draw [->] (2,0.3) -- (2,0.7); \node [below] at (2,0.4) {$x_{2}$};
\draw [->] (3,0.3) -- (3,0.7); \node [below] at (3,0.4) {$x_{3}$};
\draw [->] (4,0.3) -- (4,0.7); \node [below] at (4,0.4) {$x_{4}$};
\draw [->] (5,0.3) -- (5,0.7); \node [below] at (5,0.4) {$x_{5}$};
\node [below] at (3,0.0) {(a) RNN};

\draw [thick](8,1) circle [radius=0.3] (9,1) circle [radius=0.3] (10,1) circle [radius=0.3] (11,1) circle [radius=0.3] (12,1) circle [radius=0.3]; 
\draw [thick](8,2) circle [radius=0.3] (9,2) circle [radius=0.3] (10,2) circle [radius=0.3] (11,2) circle [radius=0.3] (12,2) circle [radius=0.3]; 
\draw [thick](8,3) circle [radius=0.3] (9,3) circle [radius=0.3] (10,3) circle [radius=0.3] (11,3) circle [radius=0.3] (12,3) circle [radius=0.3]; 
\draw (7,1) circle [radius=0.3] (7,2) circle [radius=0.3]; 
\draw (13,1) circle [radius=0.3] (13,2) circle [radius=0.3]; 
\draw [->] [thick](8, 1.3) -- (9,1.7);\draw [thick][->](12, 1.3) -- (11,1.7);\draw [thick][->](9, 2.3) -- (10,2.7);\draw [thick][->](11, 2.3) -- (10,2.7);  
\draw [dashed] [->](7, 1.3) -- (8,1.7);\draw [dashed] [->](8, 1.3) -- (8,1.7);\draw [dashed][->](9, 1.3) -- (8,1.7);\draw [dashed][->](9, 1.3) -- (9,1.7);\draw [dashed][->](10, 1.3) -- (9,1.7);\draw [dashed][->](9, 1.3) -- (10,1.7);\draw [dashed][->](10, 1.3) -- (10,1.7);\draw [dashed][->](11, 1.3) -- (10,1.7);\draw [dashed][->](10, 1.3) -- (11,1.7);\draw [dashed][->](11, 1.3) -- (11,1.7);\draw [dashed][->](11, 1.3) -- (12,1.7);\draw [dashed][->](12, 1.3) -- (12,1.7); \draw [dashed][->](13, 1.3) -- (12,1.7); 
\draw [dashed] [->](8,2.3) -- (8,2.7);\draw [dashed] [->](7,2.3) -- (8,2.7);\draw [dashed][->](9,2.3) -- (8,2.7);\draw [dashed][->](8,2.3) -- (9,2.7);\draw [dashed][->](9,2.3) -- (9,2.7);\draw [dashed][->](10,2.3) -- (9,2.7);\draw [dashed][->](10,2.3) -- (10,2.7);\draw [dashed][->](10,2.3) -- (11,2.7);\draw [dashed][->](11,2.3) -- (11,2.7);\draw [dashed][->](12,2.3) -- (11,2.7);\draw [dashed][->](11,2.3) -- (12,2.7);\draw [dashed][->](12,2.3) -- (12,2.7);\draw [dashed][->](13,2.3) -- (12,2.7);  
\draw [->] (8,0.3) -- (8,0.7); \node [below] at (8,0.4) {$x_{1}$};
\draw [->] (9,0.3) -- (9,0.7); \node [below] at (9,0.4) {$x_{2}$};
\draw [->] (10,0.3) -- (10,0.7); \node [below] at (10,0.4) {$x_{3}$};
\draw [->] (11,0.3) -- (11,0.7); \node [below] at (11,0.4) {$x_{4}$};
\draw [->] (12,0.3) -- (12,0.7); \node [below] at (12,0.4) {$x_{5}$};
\node[below] at (7,0.45) {\small padding}; \node [below] at (13,0.45) {\small padding};
\node[below] at (10,0.0) {(b) CNN};

\draw [thick](15,1) circle [radius=0.3] (16,1) circle [radius=0.3] (17,1) circle [radius=0.3] (18,1) circle [radius=0.3] (19,1) circle [radius=0.3]; 
\draw [thick](15,3) circle [radius=0.3] (16,3) circle [radius=0.3] (17,3) circle [radius=0.3] (18,3) circle [radius=0.3] (19,3) circle [radius=0.3]; 
\draw [->] [thick](15,1.3) -- (15,2.7);\draw [->] [thick](16,1.3) -- (15,2.7);\draw [->] [thick](17,1.3) -- (15,2.7);\draw [->] [thick](18,1.3) -- (15,2.7);\draw [->] [thick](19,1.3) -- (15,2.7);
\draw [->] [dashed](15,1.3) -- (16,2.7);\draw [->] [dashed](16,1.3) -- (16,2.7);\draw [->] [dashed](17,1.3) -- (16,2.7);\draw [->] [dashed](18,1.3) -- (16,2.7);\draw [->] [dashed](19,1.3) -- (16,2.7);
\draw [->] [dashed](15,1.3) -- (17,2.7);\draw [->] [dashed](16,1.3) -- (17,2.7);\draw [->] [dashed](17,1.3) -- (17,2.7);\draw [->] [dashed](18,1.3) -- (17,2.7);\draw [->] [dashed](19,1.3) -- (17,2.7);
\draw [->] [dashed](15,1.3) -- (18,2.7);\draw [->] [dashed](16,1.3) -- (18,2.7);\draw [->] [dashed](17,1.3) -- (18,2.7);\draw [->] [dashed](18,1.3) -- (18,2.7);\draw [->] [dashed](19,1.3) -- (18,2.7);
\draw [->] [dashed](15,1.3) -- (19,2.7);\draw [->] [dashed](16,1.3) -- (19,2.7);\draw [->] [dashed](17,1.3) -- (19,2.7);\draw [->] [dashed](18,1.3) -- (19,2.7);\draw [->] [dashed](19,1.3) -- (19,2.7);
\draw [->] (15,0.3) -- (15,0.7); \node [below] at (15,0.4) {$x_{1}$};
\draw [->] (16,0.3) -- (16,0.7); \node [below] at (16,0.4) {$x_{2}$};
\draw [->] (17,0.3) -- (17,0.7); \node [below] at (17,0.4) {$x_{3}$};
\draw [->] (18,0.3) -- (18,0.7); \node [below] at (18,0.4) {$x_{4}$};
\draw [->] (19,0.3) -- (19,0.7); \node [below] at (19,0.4) {$x_{5}$};
\node [below] at (17,0.0) {(c) Self-attention};
\end{tikzpicture}        
\caption{Architectures of different neural networks in NMT.}
\label{fig:nn_architectures}
\end{figure*}  

\section{Background} 
  \label{sec:background}

\subsection{NMT Architectures}
  \label{sub:nmt_architectures}

We evaluate three different NMT architectures: RNN-based 
models, CNN-based models, and Transformer-based models. 
All of them have a bipartite structure in the sense that they consist of an encoder
and a decoder.
The encoder and the decoder interact via a soft-attention mechanism 
\cite{bahdanau15joint,luong2015attention}, 
with one or multiple attention layers. 

In the following sections, $h^{l}_{i}$ is the hidden state at step 
$i$ of layer $l$, $h^{l}_{i-1}$ represents the hidden state at the previous 
step of layer $l$, $h^{l-1}_{i}$ means the hidden state at $i$ of 
$l-1$ layer, $E_{x_{i}}$ represents the embedding of $x_{i}$, and $e_{pos,i}$ 
denotes the positional embedding at position $i$. 

\subsubsection{RNN-based NMT}
  \label{ssub:rnn_based_nmt}

RNNs are stateful networks that change as new inputs are fed to them, and each
state has a direct connection only
to the previous state. Thus, the path length of any two tokens 
with a distance of $n$ in RNNs is exactly $n$. 
Figure \ref{fig:nn_architectures} (a) shows an illustration of RNNs. 

\begin{equation} \label{eq:rnn-enc-hidden}
h^{l}_{i} = h^{l-1}_{i} + f_{rnn}(h^{l-1}_{i}, h^{l}_{i-1})
\end{equation} 
In deep architectures, two adjacent layers are commonly connected with residual connections. 
In the $l$th encoder layer, $h^{l}_{i}$ is generated by 
Equation \ref{eq:rnn-enc-hidden}, where $f_{rnn}$ is the RNN 
(GRU or LSTM) function. 
In the first layer, $h^{0}_{i} = f_{rnn}(E_{x_{i}}, h^{0}_{i-1})$. 

In addition to the connection between the encoder and decoder via attention,
the initial state of the decoder is usually initialized with 
the average of the hidden states or the last hidden state of the encoder.

\subsubsection{CNN-based NMT}
  \label{ssub:cnn_based_nmt}

CNNs are hierarchical networks, in that convolution layers capture local 
correlations. The local context size depends on the size of the kernel and 
the number of layers. In order to keep the output the same length 
as the input, CNN models add padding symbols to input sequences.
Given an $L$-layer CNN with a kernel size $k$, the largest context size 
is $L(k-1)$.  
For any two tokens in a local context with a distance of $n$, 
the path between them is only $\lceil n/(k-1) \rceil$.

As Figure \ref{fig:nn_architectures} (b) shows, a 2-layer CNN with 
kernel size 3 ``sees'' an effective local context of 5 tokens. 
The path between the first token and the fifth token is only 2 convolutions. 
Since CNNs do not have a means to infer the position of elements in a
sequence, positional embeddings are introduced.

\begin{multline} \label{eq:cnn-enc-hidden}
h^{l}_{i} = h^{l-1}_{i} + f_{cnn}(W^{l}[h^{l-1}_{i-\lfloor k/2 \rfloor};...;h^{l-1}_{i+\lfloor k/2 \rfloor}]\\ + b^{l}) 
\end{multline}
The hidden state $h^{l}_{i}$ shown in Equation 
\ref{eq:cnn-enc-hidden} is related to the hidden states in the same convolution and the hidden 
state $h^{l-1}_{i}$ from the previous layer. 
$k$ denotes the kernel size 
of CNNs and $f_{cnn}$ is a non-linearity. \textit{ConvS2S} 
chooses Gated Linear Units (GLU) which can be viewed as a gated 
variation of ReLUs. $W^{l}$ are called convolutional filters. 
In the input layer, $h^{0}_{i} = E_{x_{i}} + e_{pos,i}$.

\subsubsection{Transformer-based NMT}
  \label{ssub:transformer_based_nmt}

Transformers rely heavily on self-attention networks. 
Each token is connected to any other token in the same sentence directly 
via self-attention. Moreover, Transformers feature attention networks with
multiple \textit{attention heads}. Multi-head attention is more fine-grained,
compared to conventional $1$-head attention mechanisms. 
Figure \ref{fig:nn_architectures} (c) illustrates that any two 
tokens are connected directly: the path length between the first and 
the fifth tokens is $1$. Similar to CNNs, 
positional information is also preserved in positional embeddings.

The hidden state in the Transformer encoder is calculated from all hidden states of the previous layer.
The hidden state $h^{l}_{i}$ in a self-attention network is computed as in Equation \ref{eq:self-attention}.
\begin{equation} \label{eq:self-attention}
h^{l}_{i} = h^{l-1}_{i} + f(\text{self-attention}(h^{l-1}_{i}))
\end{equation} 
where $f$ represents a feed-forward network with ReLU as the activation function and layer normalization.
In the input layer, $h^{0}_{i} = E_{x_{i}} + e_{pos,i}$. 
The decoder additionally has a multi-head attention over the encoder hidden states.

\subsection{Contrastive Evaluation of Machine Translation} 
  \label{sub:contrastive_evaluation}

Since we evaluate different NMT architectures explicitly on subject-verb agreement and WSD
(both happen implicitly during machine translation), BLEU as a measure of overall translation
quality is not helpful. In order to conduct these targeted evaluations, we use contrastive test sets.

Sets of contrastive translations can be used to analyze specific types of errors.
Human reference translations are paired with one or more contrastive variants, 
where a specific type of error is introduced automatically.

The evaluation procedure then exploits the fact that NMT models are conditional language models. By virtue of this, given any source sentence $S$ and target sentence $T$, any NMT
model can assign to them a probability $P(T|S)$. If a model assigns a higher score to the correct target sentence than to a contrastive variant that contains an error, we consider it a correct decision.
The accuracy of a model on such a test set is simply the percentage of cases where the correct target sentence
is scored higher than all contrastive variants.

Contrastive evaluation tests the sensitivity of NMT models to specific translation errors. 
The contrastive examples are designed to capture specific translation errors rather 
than evaluating the global quality of NMT models. 
Although they do not replace metrics such as BLEU, they give further insights into the performance of models, on specific linguistic phenomena. 

\subsubsection{\textit{Lingeval97}}

\textit{Lingeval97} has over 97,000 English$\rightarrow$German contrastive 
translation pairs featuring different linguistic phenomena, including 
subject-verb agreement, noun phrase agreement, separable verb-particle 
constructions, transliterations and polarity.  
In this paper, we are interested in evaluating the performance on 
long-range dependencies. Thus, we focus on the subject-verb agreement 
category which consists of 35,105 instances. 

In German, verbs must agree with their subjects in both grammatical number 
and person. Therefore, in a contrastive translation, the grammatical number of a verb is swapped.
Table \ref{table-sv-example} gives an example. 

\begin{table}[htbp]
\begin{center}
\begin{tabular}{ll}
English: & [...] plan will be approved\\ 
 German: & [...] \textbf{Plan} verabschiedet \textbf{wird}\\ 
Contrast:& [...] \textbf{Plan} verabschiedet \textbf{werden}\\ 

\end{tabular}
\caption{\label{table-sv-example} An example of a contrastive pair in the 
subject-verb agreement category. }
\end{center}
\end{table}

\subsubsection{\textit{ContraWSD}}

In \textit{ContraWSD}, given an ambiguous word in the source sentence, 
the correct translation is replaced by another meaning of the ambiguous word 
which is incorrect. 
For example, in a case where the English word \textit{line} is the correct translation of the German source word \textit{Schlange}, \textit{ContraWSD} replaces \textit{line} with the other translations of \textit{Schlange}, such as 
\textit{snake}, \textit{serpent}, to generate contrastive translations. 

For German$\rightarrow$English, \textit{ContraWSD} contains 84 different 
German word senses. 
It has 7,200 German$\rightarrow$English lexical ambiguities, each lexical 
ambiguity instance has 3.5 contrastive translations on average. 
For German$\rightarrow$French, it consists of 71 different German word senses. 
There are 6,700 German$\rightarrow$French lexical ambiguities, with an 
average of 2.2 contrastive translations each lexical ambiguity instance. 
All the ambiguous words are nouns so that the disambiguation is 
not possible simply based on syntactic context.

\section{Subject-verb Agreement}
  \label{sec:subect-verb_agreement}

The subject-verb agreement task is the most popular choice for 
evaluating the ability to capture long-range dependencies and
 has been used in many studies \cite{Linzen2016assessing,bernardy2017using,sennrich2017grammatical,Tran2018recurrent}. 
Thus, we also use this task to evaluate different NMT architectures 
on long-range dependencies. 

\subsection{Experimental Settings}
  \label{sub:experimental_settings}

Different architectures are hard to compare fairly because many factors 
affect performance. We aim to create a level playing field for the comparison 
by training with the same toolkit, \textit{Sockeye} \cite{Hieber2017sockeye} 
which is based on MXNet \cite{chen2015mxnet}. 
In addition, different hyperparameters and training techniques (such as label smoothing or layer normalization) have been found to affect the performance \cite{chen2018both}. 
We apply the same hyperparameters and techniques for all 
architectures except the parameters of each specific architecture. 
Since the best hyperparameters for different architectures may be diverse, 
we verify our hyperparameter choice by comparing our results to those published previously. 
Our models achieve similar performance to that reported by \newcite{Hieber2017sockeye} 
with the best available settings.
In addition, we extend \textit{Sockeye} with an interface that enables
scoring of existing translations,
which is required for contrastive evaluation. 

All the models are trained with 2 GPUs. 
During training, each mini-batch contains 4096 tokens. 
A model checkpoint is saved every 4,000 updates. 
We use \textit{Adam} \cite{Kingma2014AdamAM} as the optimizer. 
The initial learning rate is set to 0.0002. 
If the performance on the validation set has not improved for 8 
checkpoints, the learning rate is multiplied by 0.7. 
We set the early stopping patience to 32 checkpoints. 
All the neural networks have 8 layers. For \textit{RNNS2S}, the 
encoder has 1 bi-directional LSTM and 6 stacked uni-directional LSTMs, 
and the decoder is a stack of 8 uni-directional LSTMs. 
The size of embeddings and hidden states is 512.
We apply layer normalization and label smoothing (0.1) in all models.
We tie the source and target embeddings. 
The dropout rate of embeddings and Transformer blocks is set to 0.1. 
The dropout rate of RNNs and CNNs is 0.2. 
The kernel size of CNNs is 3. 
Transformers have an 8-head attention mechanism. 

To test the robustness of our findings, we also test a different style 
of RNN architecture, from a different toolkit. 
We evaluate bi-deep transitional RNNs \cite{miceli2017deepRNN} 
which are state-of-art RNNs in machine translation. 
We use the bi-deep RNN-based model (\textit{RNN-bideep}) implemented 
in \textit{Marian} \cite{marcin2018Marian}. 
Different from the previous settings, we use the Adam optimizer with 
$\beta_{1}=0.9$, $\beta_{2}=0.98$, $\epsilon=10^{-9}$. The initial 
learning rate is 0.0003. We tie target embeddings and output 
embeddings. Both the encoder and decoder have 4 layers of LSTM units, only the encoder
layers are bi-directional. LSTM units consist of several cells (deep transition): 4 in
the first layer of the decoder, 2 cells everywhere else.

We use training data from the WMT17 shared 
task.\footnote{\url{http://www.statmt.org/wmt17/translation-task.html}} 
We use \textit{newstest2013} as the validation set, and use 
\textit{newstest2014} and \textit{newstest2017} as the test sets. 
All BLEU scores are computed with \textit{SacreBLEU} \cite{post2018sacre}. 
There are about 5.9 million sentence pairs in the training set after 
preprocessing with Moses scripts. 
We learn a joint BPE model with 32,000 subword units \cite{sennrich16sub}. 
We employ the model that has the best perplexity on the validation set 
for the evaluation.

\subsection{Overall Results}
  \label{sub:overall_results}

Table \ref{table-bleu-sv} reports the BLEU scores on \textit{newstest2014} 
and \textit{newstest2017}, the perplexity on the validation set, 
and the accuracy on long-range dependencies.\footnote{We report average accuracy on instances where the distance between subject and verb is longer than 10 words.}
\textit{Transformer} achieves the highest accuracy on this task and the highest BLEU scores on both \textit{newstest2014} 
and \textit{newstest2017}. 
Compared to \textit{RNNS2S}, \textit{ConvS2S} has slightly better results 
regarding BLEU scores, but a much lower accuracy on long-range dependencies. 
The \textit{RNN-bideep} model achieves distinctly better BLEU scores and a higher 
accuracy on long-range dependencies. However, it still cannot outperform Transformers on
any of the tasks.

\begin{table}[htbp]
\begin{center}
\begin{tabular}{|l|c|c|c|c|}
\hline \bf Model & 2014& 2017 &PPL&Acc(\%)\\ 
\hline \textit{RNNS2S} & 23.3& 25.1&6.1&95.1\\
\textit{ConvS2S} &23.9 & 25.2 &7.0&84.9\\
\textit{Transformer} &\textbf{26.7}  &\textbf{27.5}&\textbf{4.5}&\textbf{97.1}\\
\hline \textit{RNN-bideep} & 24.7& 26.1&5.7&96.3\\
 %\textit{Trans-Marian} &26.09  &27.01&4.86&97.48\\
\hline
\end{tabular}
\caption{\label{table-bleu-sv} The results of different NMT models, 
including the BLEU scores on \textit{newstest2014} and \textit{newstest2017}, 
the perplexity on the validation set, and the accuracy of long-range dependencies.}
\end{center}
\end{table}

\begin{figure}[htbp]
\centering
        \includegraphics[totalheight=5.5cm]{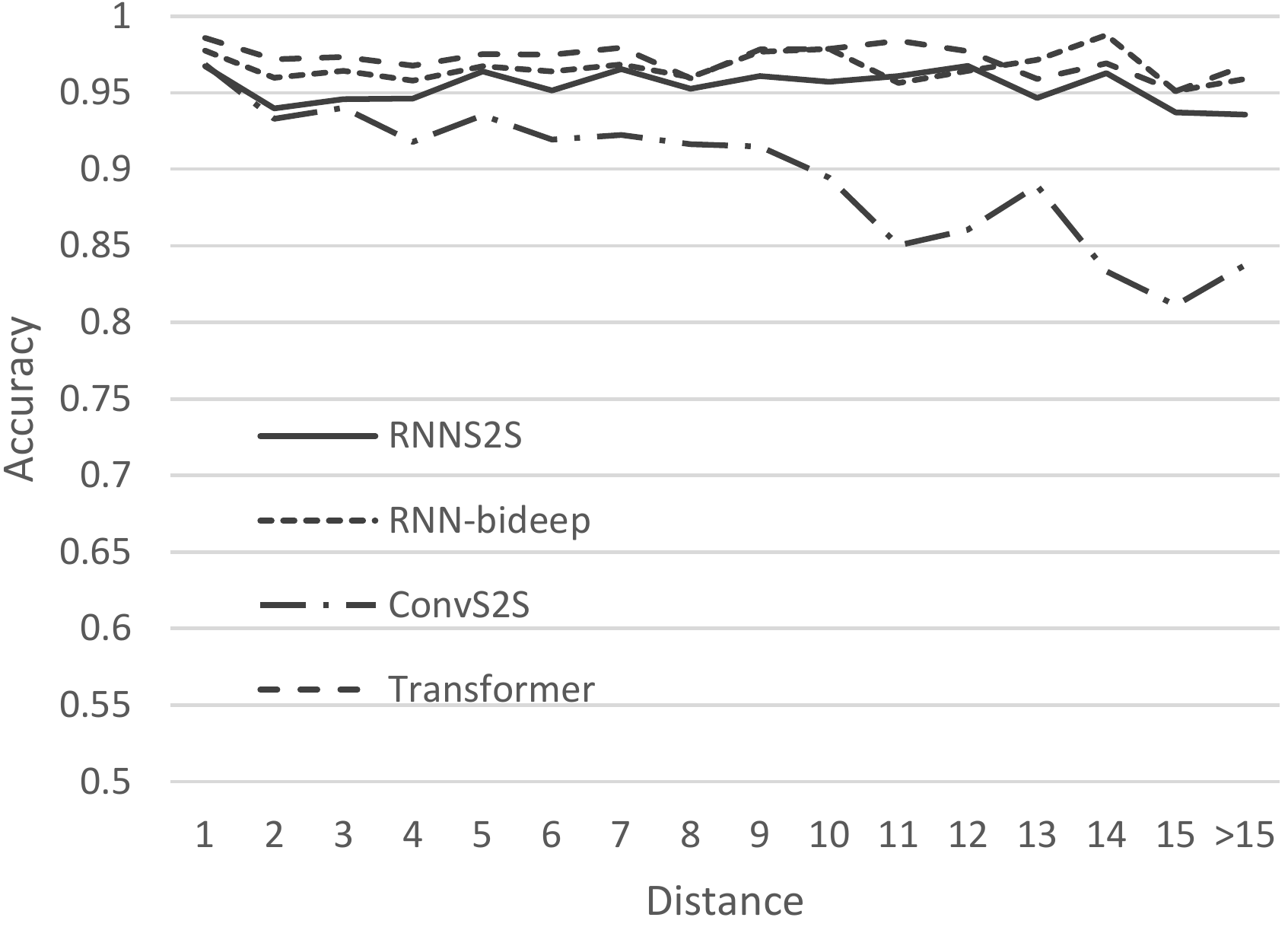}
    \caption{Accuracy of different NMT models on the subject-verb agreement task.}
    \label{fig:sv-overall}
\end{figure}

\noindent
Figure \ref{fig:sv-overall} shows the performance of different 
architectures on the subject-verb agreement task. 
It is evident that \textit{Transformer}, \textit{RNNS2S}, and 
\textit{RNN-bideep} perform much better than \textit{ConvS2S} on 
long-range dependencies. 
However, \textit{Transformer}, \textit{RNNS2S}, and \textit{RNN-bideep} 
are all robust over long distances. 
\textit{Transformer} outperforms \textit{RNN-bideep} for distances 11-12, 
but \textit{RNN-bideep} performs equally or better for distance 13 or higher. 
Thus, we cannot conclude that Transformer models are particularly stronger than 
RNN models for long distances, despite achieving higher average accuracy on distances above 10.

\subsubsection{CNNs} 
  \label{ssub:cnns}

Theoretically, the performance of CNNs will drop when the distance between 
the subject and the verb exceeds the local context size.
However, \textit{ConvS2S} is also clearly worse than \textit{RNNS2S} for subject-verb agreement within the local context size.

\begin{figure}[htbp]
\centering
        \includegraphics[totalheight=5.5cm]{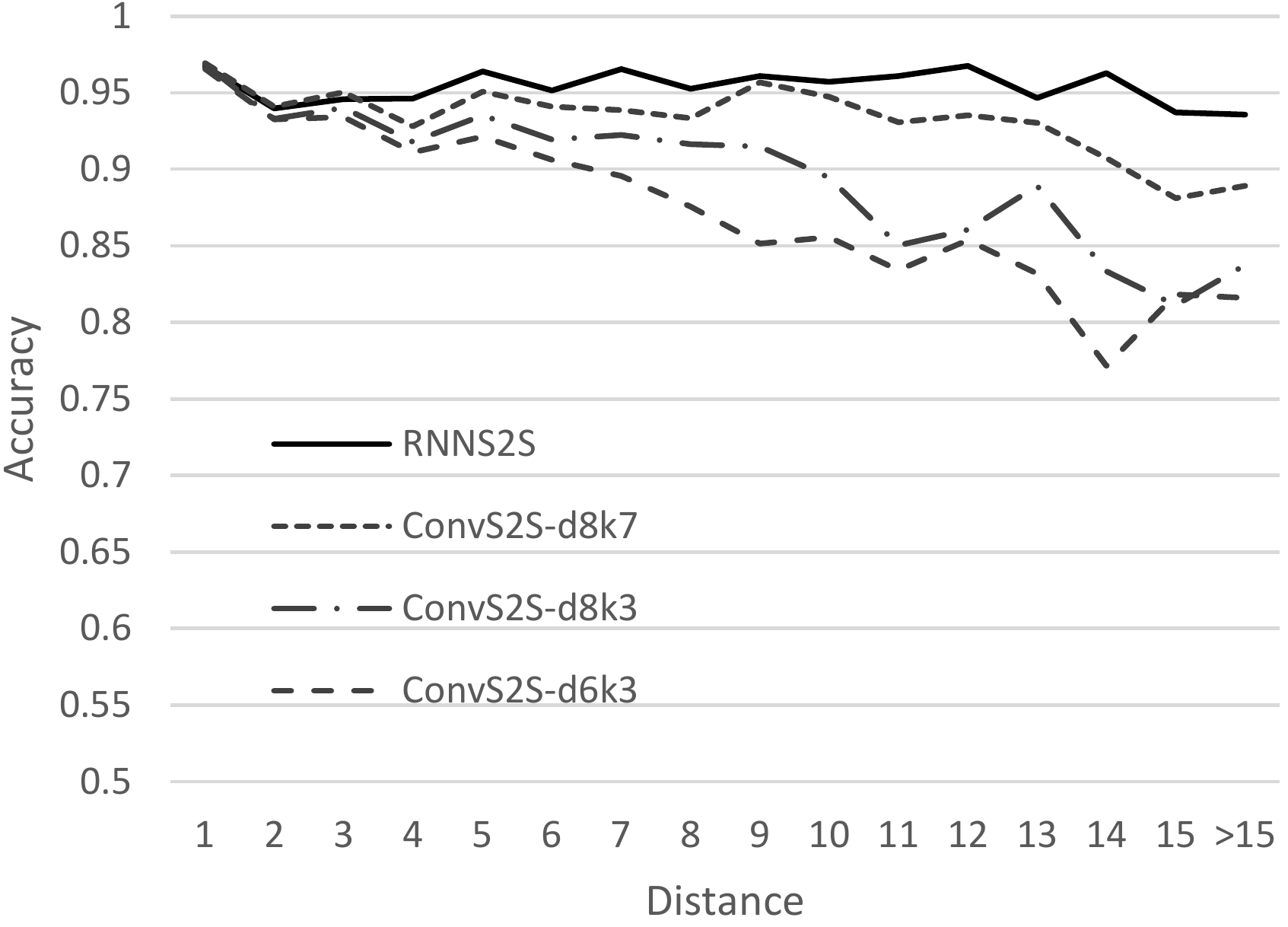}
    \caption{Results of \textit{ConvS2S} models and the \textit{RNNS2S} model 
at different distances. }
    \label{fig:sv-cnn}
\end{figure}

In order to explore how the ability of \textit{ConvS2S} to capture long-range dependencies 
depends on the local context size, we train additional systems, varying the number of layers
and kernel size. Table \ref{table-cnn} shows the performance of different \textit{ConvS2S} models. 
Figure \ref{fig:sv-cnn} displays the performance of two 8-layer CNNs with 
kernel size 3 and 7, a 6-layer CNN with kernel size 3, and \textit{RNNS2S}. 
The results indicate that the accuracy increases when the local context 
size becomes larger, but the BLEU score does not.
Moreover, \textit{ConvS2S} is still not as good as \textit{RNNS2S} for subject-verb agreement.

\begin{table}[htbp]
\begin{center}
\begin{tabular}{|c|c|c|c|c|c|}
\hline Layer & K &Ctx& 2014 & 2017& Acc(\%) \\
\hline 4&3&\phantom{0}8&22.9&24.2&81.1 \\
\hline 6&3&12&23.6&25.0&82.5 \\
\hline 8&3&16&\textbf{23.9}& \textbf{25.2}&84.9\\
\hline 8&5&32&23.5&24.7&89.7\\
\hline 8&7&48&23.3&24.6&\textbf{91.3}\\
\hline
\end{tabular}
\caption{\label{table-cnn} The performance of \textit{ConvS2S} with 
different settings. \textit{K} means the kernel size. The \textit{ctx} 
column is the theoretical largest local context size in the masked decoder.}
\end{center}
\end{table}

\noindent
Regarding the explanation for the poor performance of \textit{ConvS2S}, 
we identify the limited context size as a major problem. 
One assumption to explain the remaining difference is that, 
scale invariance of CNNs is relatively poor \cite{xu2014scale}. 
Scale-invariance is important in NLP, where the distance between arguments is flexible,  
and current recurrent or attentional architectures are better suited to handle this variance.

Our empirical results do not confirm the theoretical arguments in \newcite{gehring2017convolutional}
that CNNs can capture long-range dependencies better with a shorter path. 
The BLEU score does not correlate well with the targeted evaluation of 
long-range distance interactions. 
This is due to the locality of BLEU, which only measures 
on the level of n-grams, but it may also indicate that there are other 
trade-offs between the modeling of different phenomena depending on 
hyperparameters. If we aim to get better performance 
on long-range dependencies, we can take this into account 
when optimizing hyperparameters.

\subsubsection{RNNs vs. \textit{Transformer}}
  \label{ssub:rnns_vs_textit_transformer}

Even though \textit{Transformer} achieves much better BLEU scores 
than \textit{RNNS2S} and \textit{RNN-bideep}, the accuracies of these 
architectures on long-range dependencies are close to each other 
in Figure \ref{fig:sv-overall}. 

Our experimental result contrasts with the result from 
\newcite{Tran2018recurrent}. They find that Transformers perform
worse than LSTMs on the subject-verb agreement task, especially 
when the distance between the subject and the verb becomes longer. 
We perform several experiments to analyze this discrepancy 
with \newcite{Tran2018recurrent}.

\begin{figure}[htbp]
\centering
        \includegraphics[totalheight=5.5cm]{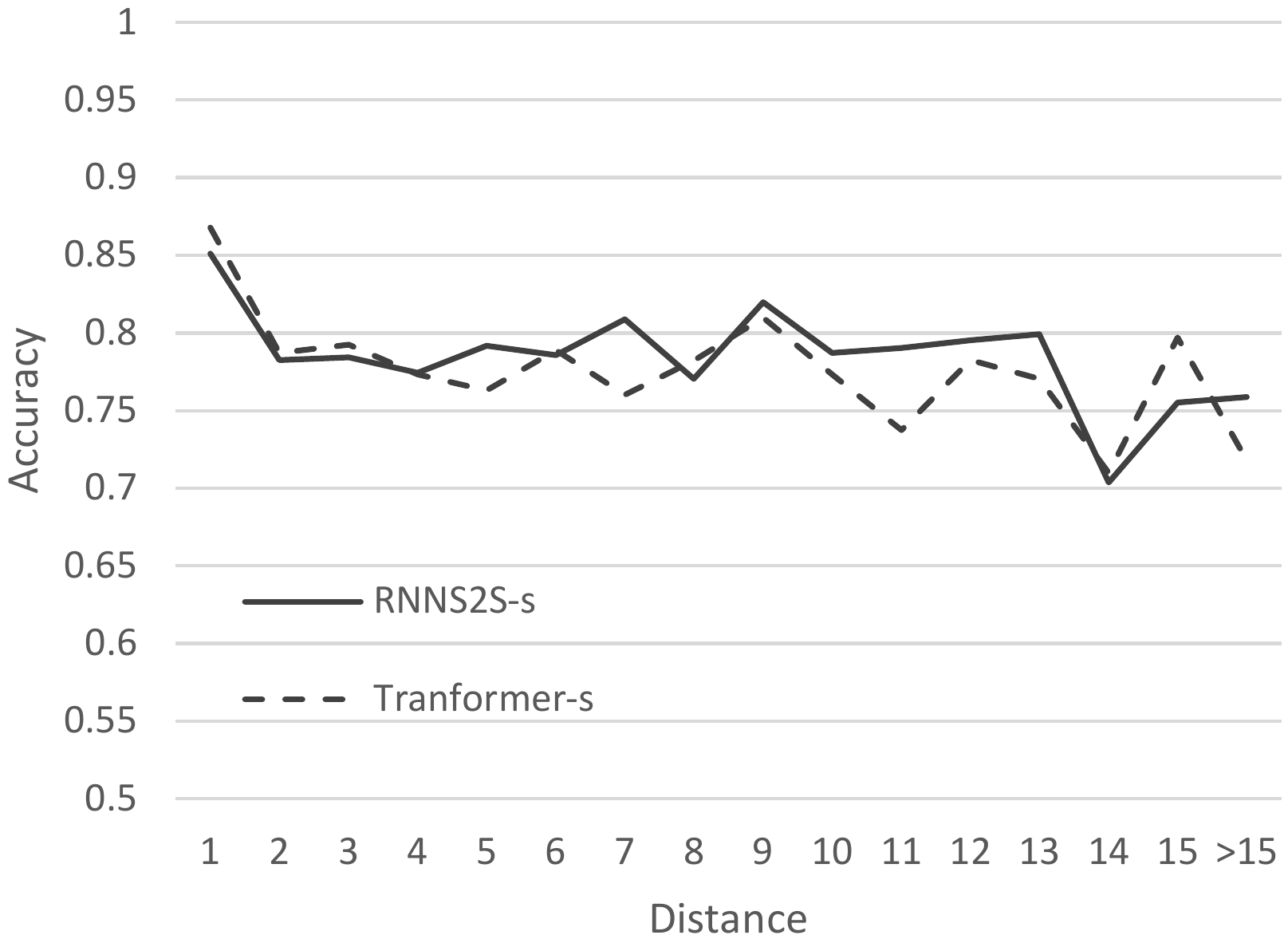}
    \caption{Results of a Transformer and \textit{RNNS2S} model trained on a small dataset. }
    \label{fig:sv-trans-rnn}
\end{figure}

A first hypothesis is that this is caused by the amount of training data, 
since we used much larger datasets than \newcite{Tran2018recurrent}.
We retrain all the models with a small amount of training data similar to 
the amount used by \newcite{Tran2018recurrent}, about 135K sentence pairs. 
The other training settings are the same as in 
Section \ref{sub:experimental_settings}. 
We do not see the expected degradation of \textit{Transformer-s}, 
compared to \textit{RNNS2S-s} (see Figure \ref{fig:sv-trans-rnn}). 
In Table \ref{table-res-small}, the performance of \textit{RNNS2S-s} 
and \textit{Transformer-s} is similar, including the BLEU scores 
on \textit{newstest2014, newstest2017}, the perplexity on the validation set, 
and the accuracy on the long-range dependencies. 

\noindent
A second hypothesis is that the experimental settings lead to the different results. 
In order to investigate this, we do not only use a small training set, 
but also replicate the experimental settings of \newcite{Tran2018recurrent}. 
The main changes are neural network layers (8$\rightarrow$4); 
embedding size (512$\rightarrow$128); multi-head size (8$\rightarrow$2); 
dropout rate (0.1$\rightarrow$0.2); 
checkpoint save frequency (4,000$\rightarrow$1,000), and 
initial learning rate (0.0002$\rightarrow$0.001). 

\begin{table}[htbp]
\begin{center}
\begin{tabular}{|l|c|c|c|c|}
\hline Model & 2014 & 2017 & PPL&Acc(\%)\\
\hline \textit{RNNS2S-s}&7.3&\phantom{0}7.8&47.8&77.3\\
\textit{Trans-s}&7.2& \phantom{0}8.0&44.6&74.6\\
\hline \textit{RNNS2S-re}&9.2&10.5&39.2&77.7\\
\textit{Trans-re-h2}&9.6&10.7&36.9&71.9\\
\textit{Trans-re-h4}&9.5&11.9&35.8&73.8\\
\textit{Trans-re-h8}&9.4&10.4&36.0&75.3\\
\hline
\end{tabular}
\caption{\label{table-res-small} The results of different models with small 
training data and replicate settings. 
\textit{Trans} is short for Transformer.
Models with the suffix ``\textit{-s}'' are models trained with small data set. 
Models with the suffix ``\textit{-re}'' are models trained with replicate settings. 
``\textit{h2, h4, h8}'' indicates the number of attention heads for Transformer models. }
\end{center}
\end{table}

\begin{figure}[htbp]
\centering
        \includegraphics[totalheight=5.5cm]{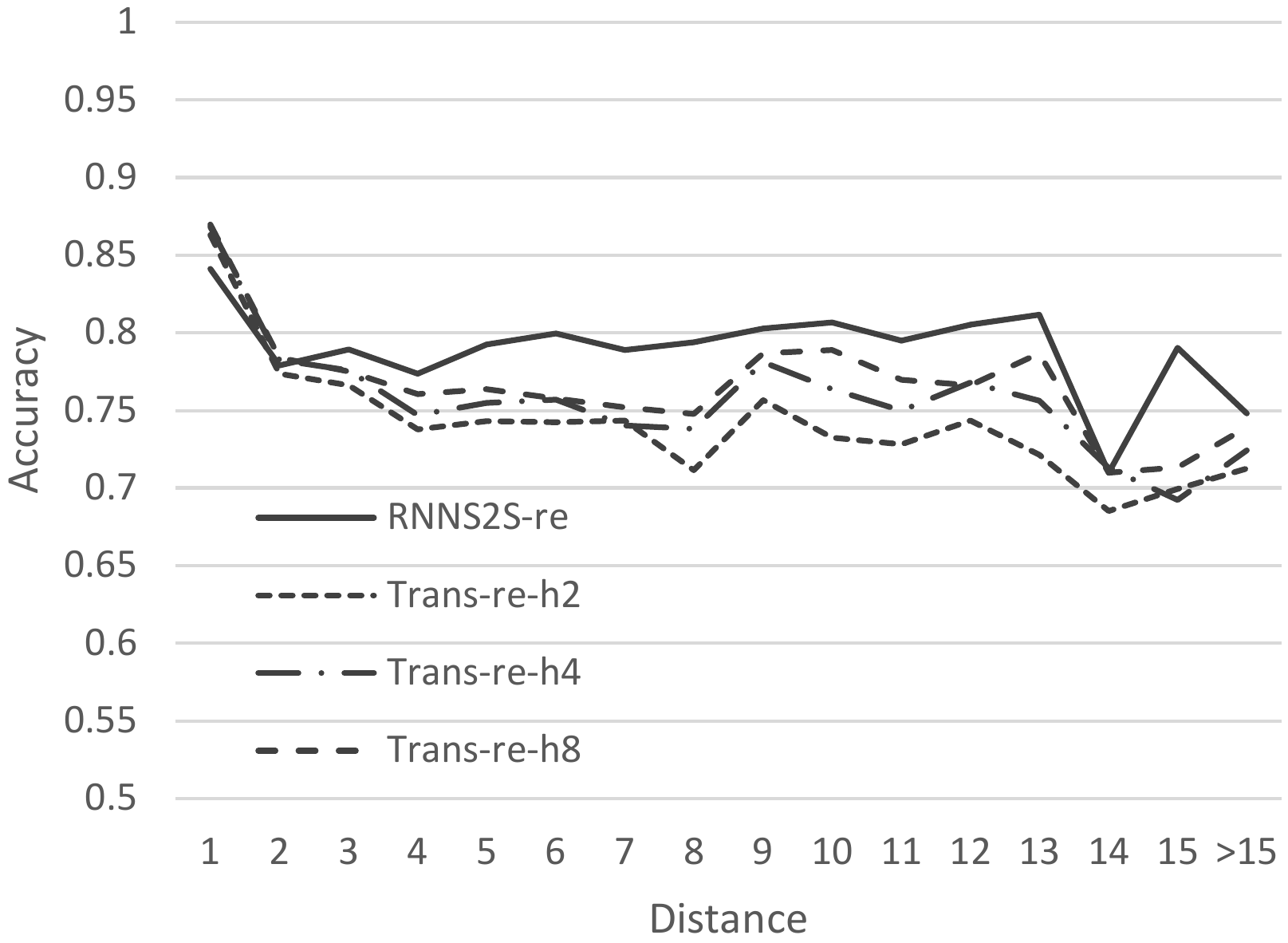}
    \caption{Results of the models with replicate settings, varying the number of attention heads for the Transformer models.}
    \label{fig:sv-rnn-trans-head}
\end{figure}  

\noindent
In the end, we get a result that is similar to \newcite{Tran2018recurrent}. 
In Figure \ref{fig:sv-rnn-trans-head}, \textit{Transformer-re-h2} performs 
clearly worse than \textit{RNNS2S-re} on long-range dependencies. 
By increasing the number of heads in multi-head attention, subject-verb accuracy over long distances can be improved substantially, even though it remains below that of \textit{RNNS2S-re}.
Also, the effect on BLEU is small.

Our results suggest that the importance of multi-head attention with a large number of heads is larger than BLEU would suggest, especially for the modeling of long-distance phenomena, since multi-head attention provides a way for the model to attend to both local and distant context, whereas distant context may be overshadowed by local context in an attention mechanism with a single or few heads.

Although our study is not a replication of \newcite{Tran2018recurrent}, 
who work on a different task and a different test set, our results 
do suggest an alternative interpretation of their findings, namely that 
the poor performance of the Transformer in their experiments 
is due to hyperparameter choice.  
Rather than concluding that RNNs are superior to Transformers for the modeling of long-range dependency phenomena, we find that the number of heads in multi-head attention affects the ability of Transformers to model long-range dependencies in subject-verb agreement. 

\begin{table*}[ht]
\begin{center}
\begin{tabular}{|l|cccc|ccc|}
\hline \multirow{2}{*}{Model} & \multicolumn{4}{|c|}{DE$\rightarrow$EN} & \multicolumn{3}{c|}{DE$\rightarrow$FR} \\ 
\cline{2-8}  & PPL & 2014 & 2017 & Acc(\%) & PPL & 2012 & Acc(\%)\\ 
\hline \textit{RNNS2S} &5.7&29.1 &30.1& 84.0 & 7.06&16.4 &72.2 \\ 
\textit{ConvS2S} & 6.3&29.1&30.4&82.3 & 7.93&16.8&72.7  \\ 
\textit{Transformer} & \textbf{4.3}&\textbf{32.7}&33.7 &\textbf{90.3} & \textbf{4.9}&\textbf{18.7}& \textbf{76.7} \\ 
\hline \textit{uedin-wmt17} & --& --&\textbf{35.1}& 87.9 &-- &-- &-- \\ 
\hline \textit{TransRNN} & 5.2& 30.5&31.9&86.1& 6.3&17.6 &74.2 \\ 
\hline
\end{tabular}
\caption{\label{table-WSD} The results of different architectures on 
\textit{newstest} sets and \textit{ContraWSD}. 
\textit{PPL} is the perplexity on the validation set. 
\textit{Acc} means accuracy on the test set. 
}\end{center}
\end{table*}

\section{WSD}
  \label{sec:word_sense_disambiguation}

Our experimental results on the subject-verb agreement task demonstrate 
that CNNs and Transformer are not better at capturing 
long-range dependencies compared to RNNs, even though the paths in CNNs 
and Transformers are shorter. 
This finding is not in accord with the theoretical argument in both 
\newcite{gehring2017convolutional} and \newcite{vaswani2017Attention}. 
However, these architectures perform well empirically according to BLEU. 
Thus, we further evaluate these architectures on WSD, to test our 
hypothesis that non-recurrent architectures are better at extracting semantic features.

\subsection{Experimental settings}
  \label{sub:experimental_settings1}

We evaluate all architectures on \textit{ContraWSD} on both DE$\rightarrow$EN and 
DE$\rightarrow$FR. 
We reuse the parameter settings in Section \ref{sub:experimental_settings}, 
except that: the initial learning rate of \textit{ConvS2S} is reduced 
from 0.0003 to 0.0002 in DE$\rightarrow$EN; the checkpoint saving frequency 
is changed from 4,000 to 1,000 in DE$\rightarrow$FR because of the training data size.

For DE$\rightarrow$EN, the training set, validation set, and test set 
are the same as the other direction EN$\rightarrow$DE. 
For DE$\rightarrow$FR, we use around 2.1 million sentence pairs from 
Europarl (v7) \cite{tiedman2012opus}\footnote{\url{http://opus.nlpl.eu/Europarl.php}} 
and News Commentary (v11) cleaned by 
\newcite{rios2017improving}\footnote{\url{http://data.statmt.org/ContraWSD/}} 
as our training set. 
We use \textit{newstest2013} as the evaluation set, and use 
\textit{newstest2012} as the test set. 
All the data is preprocessed with Moses scripts. 

In addition, we also compare to the best result reported for 
DE$\rightarrow$EN, achieved by \textit{uedin-wmt17} 
\cite{sennrich2017WMT}, which is an ensemble of 4 different models 
and reranked with right-to-left models.\footnote{\url{https://github.com/a-rios/ContraWSD/tree/master/baselines}} \textit{uedin-wmt17} is based on the bi-deep RNNs \cite{miceli2017deepRNN} that we mentioned before. 
To the original 5.9 million sentence pairs in the training set, they add 10 
million synthetic pairs with back-translation.

\subsection{Overall Results}
\label{sub:overall_results1}

Table \ref{table-WSD} gives the performance of all the architectures, 
including the perplexity on validation sets, the BLEU scores on 
\textit{newstest}, and the accuracy on \textit{ContraWSD}. 
Transformers distinctly outperform \textit{RNNS2S} 
and \textit{ConvS2S} models on DE$\rightarrow$EN and 
DE$\rightarrow$FR. Moreover, the Transformer model on 
DE$\rightarrow$EN also achieves higher accuracy than \textit{uedin-wmt17}, although the BLEU score 
on \textit{newstest2017} is 1.4 lower than \textit{uedin-wmt17}.
We attribute this discrepancy between BLEU and WSD performance to the use of synthetic news training data in \textit{uedin-wmt17},
which causes a large boost in BLEU due to better domain adaptation to \textit{newstest}, but which is less helpful for ContraWSD, whose test set is drawn from a variety of domains.

For DE$\rightarrow$EN, \textit{RNNS2S} and \textit{ConvS2S} have the same 
BLEU score on \textit{newstest2014}, \textit{ConvS2S} has a higher score 
on \textit{newstest2017}. However, the WSD accuracy of \textit{ConvS2S} 
is 1.7\% lower than \textit{RNNS2S}. 
For DE$\rightarrow$FR, \textit{ConvS2S} achieves slightly better results 
on both BLEU scores and accuracy than \textit{RNNS2S}. 

The Transformer model strongly outperforms the other architectures 
on this WSD task, with a gap of 4--8 percentage points. This affirms our hypothesis that Transformers are 
strong semantic features extractors.

\subsection{Hybrid Encoder-Decoder Model} 
  \label{sub:hybrid_encoder_decoder_model}

In recent work, \newcite{chen2018both} find that hybrid architectures 
with a Transformer encoder and an RNN decoder can outperform 
a pure Transformer model. They speculate that the 
Transformer encoder is better at encoding or extracting features 
than the RNN encoder, whereas the RNN is better at conditional language modeling. 

For WSD, it is unclear whether
the most important component is the encoder, the decoder, or both.
Following the hypothesis that Transformer encoders excel as semantic feature extractors,
we train a hybrid encoder-decoder model (\textit{TransRNN}) with a 
Transformer encoder and an RNN decoder. 

The results (in Table \ref{table-WSD}) show that \textit{TransRNN} performs  better than \textit{RNNS2S},
but worse than the pure Transformer, both in terms of BLEU and WSD accuracy.
This indicates that WSD is not only done in 
the encoder, but that the decoder also affects WSD performance. 
We note that \newcite{chen2018both,Domhan2018how} introduce the techniques 
in Transformers into RNN-based models, with reportedly higher BLEU.
Thus, it would be interesting to see if the same result holds true with their architectures.

\section{Post-publication Experiments} 
\label{sec:further_experiments}

\begin{table*}[ht]
\begin{center}
\begin{tabular}{|l|cccc|ccc|}
\hline \multirow{2}{*}{Model} & \multicolumn{4}{|c|}{DE$\rightarrow$EN} & \multicolumn{3}{c|}{DE$\rightarrow$FR} \\ 
\cline{2-8}  & PPL & 2014 & 2017 & Acc(\%) & PPL & 2012 & Acc(\%)\\ 
\hline \textit{RNNS2S} &4.7&31.1 &32.2& 88.1 & 5.6&17.7 &75.9 \\ 
\textit{ConvS2S} & 5.0&30.9&32.2&87.2 & 5.9&17.9&74.7  \\ 
\textit{Transformer} & \textbf{4.5}&\textbf{32.0}&\textbf{33.3} &\textbf{89.1} & \textbf{5.2}&\textbf{18.4}& \textbf{76.8} \\ 
\hline \textit{TransRNN} & 4.7& 31.6&32.9&87.8&5.4 &18.3 &76.5 \\ 
\hline
\end{tabular}
\caption{\label{table-further-WSD} Post-publication results of different architectures on 
\textit{newstest} sets and \textit{ContraWSD}. 
\textit{PPL} is the perplexity on the validation set. 
\textit{Acc} means accuracy on the test set. 
}\end{center}
\end{table*}

We here present a number of further experiments with different configurations and implementations, performed after publication to test the robustness of our claims.

\subsection{Pre-trained \textit{Fairseq} CNN Model}

The \textit{ConvS2S} models underperform \textit{RNNS2S} and Transformer on the subject-verb agreement task.
To address the question whether these results can be attributed to a misconfiguration or implementation difference in \textit{Sockeye},
we also obtained results with a pre-trained model released by \cite{gehring2017convolutional} and trained with \textit{Fairseq}\footnote{\url{https://github.com/pytorch/fairseq}}.
This pre-trained model also uses the WMT17 data set for training. 

\noindent
Table \ref{table-further-cnn} shows the model differences and performance. 
The pre-trained \textit{Fairseq} model has 15 layers, which is much deeper than the \textit{Sockeye} models that we trained. 
It achieves higher BLEU score on \textit{newstest2014}, and higher accuracy on modeling 
long-range dependencies, than the 8-layer \textit{Sockeye} models that we trained.
However, it still lags behind \textit{RNNS2S} and Transformer on the subject-verb agreement task.
%We note that our \textit{Sockeye-2} model with a larger kernel size performs close to \textit{Fairseq} in subject-verb-agreement accuracy.

\begin{table}[htbp]
\begin{center}
\begin{tabular}{|c|c|c|c|c|}
\hline Model &Layer & K & \textit{2014} & Accuracy(\%) \\
\hline \textit{Sockeye-1} &8&3 &23.9& 84.9  \\ 
\hline \textit{Sockeye-2} &8&7 &23.3& 91.3  \\ 
\hline \textit{Fairseq}& \textbf{15}&3&25.2&\textbf{92.7}\\ 
\hline
\end{tabular}
\caption{\label{table-further-cnn} The performance of CNN models trained by different toolkits. \textit{K} is the kernel size of CNN.}
\end{center}
\end{table}

\subsection{Reducing Model Differences}

The difference between recurrent, convolutional, and self-attentional architectures is not the only difference between 
the \textit{RNNS2S}, \textit{ConvS2S}, and \textit{Transformer} networks that we tested. For example, 
Transformer has multiple attention layers, multi-head attention, 
residual feed-forward layers, etc. These modules may affect NMT models on  
capturing long-range dependencies and extracting semantic features. 

\newcite{Domhan2018how} applies these advanced techniques of Transformer models to 
both RNN and CNN models in \textit{Sockeye}, minimizing the architectural difference between them.\footnote{\url{https://github.com/awslabs/sockeye/tree/acl18}} 
We reuse his configurations to train minimally different RNN, CNN and Transformer models.
All models have 6-layer encoders and decoders.

\subsubsection{Subject-verb agreement}

Table \ref{table-further-sv} gives the results of retrained models. 
Compared to the original results in Table \ref{table-bleu-sv},
we find that these configurations have a large positive effect on BLEU and perplexity of \textit{RNNS2S} and \textit{ConvS2S},
but the effect on subject-verb-agreement over long distances is relatively small.
These result further confirm our experimental results in Section \ref{sec:subect-verb_agreement} that non-recurrent neural networks are not superior to RNNs in capturing long-range dependencies. 

\begin{table}[htbp]
\begin{center}
\begin{tabular}{|l|c|c|c|c|}
\hline  \bf Model & 2014& 2017 &PPL&Acc(\%)\\ 
\hline \textit{RNNS2S} & 25.6& 26.5&4.9&\textbf{96.9}\\
\textit{ConvS2S} &25.4 & 26.6 &5.4&85.0\\
\textit{Transformer} &\textbf{26.1}  &\textbf{27.4}&\textbf{4.7}&96.6\\
\hline
\end{tabular}
\caption{\label{table-further-sv} Post-publication results, 
including BLEU on \textit{newstest2014} and \textit{newstest2017}, 
perplexity on the validation set, and accuracy of long-range dependencies.}
\end{center}
\end{table}

\subsubsection{WSD}

The performance of retrained models on WSD task is shown in Table 
\ref{table-further-WSD}. Compared to the original results in Table \ref{table-WSD},
the performance gap between Transformer models and the other models is 
getting smaller across all metrics (BLEU, perplexity, and WSD accuracy), although Transformer still performs best.
This implies that some of the strong performance of the Transformer architecture for WSD is attributable
to architecture choices such as multi-head attention, layer normalization, and upscaling feed-forward layers in each block.
Nevertheless, the retrained \textit{RNNS2S} and \textit{ConvS2S} models are still 
not as good as the retrained Transformer models, 
so these results also further confirm our results in Section \ref{sec:word_sense_disambiguation}.

\section{Conclusion} 
  \label{sec:conclusion}

In this paper, we evaluate three popular NMT architectures, 
\textit{RNNS2S}, \textit{ConvS2S}, and Transformers, 
on subject-verb agreement and WSD by scoring 
contrastive translation pairs. 

We test the theoretical claims that shorter path lengths make models
better capture long-range dependencies. Our experimental results show that: 
\begin{itemize}
  \item There is no evidence that CNNs and Transformers, which have 
    shorter paths through networks, are empirically superior to RNNs in 
    modeling subject-verb agreement over long distances. 
  \item The number of heads in multi-head attention affects
    the ability of a Transformer to model long-range dependencies 
    in the subject-verb agreement task. 
  \item Transformer models excel at another task, WSD, 
  compared to the CNN and RNN architectures we tested. 
\end{itemize}

\noindent
Lastly, our findings suggest that assessing the performance
of NMT architectures means finding their inherent trade-offs,
rather than simply computing their overall BLEU score. A clear understanding of
those strengths and weaknesses is important to guide further work. Specifically,
given the idiosyncratic limitations of recurrent and self-attentional models,
combining them is an exciting line of research.
The apparent weakness of CNN architectures on long-distance phenomena is also a problem worth tackling,
and we can find inspiration from related work in computer vision \cite{xu2014scale}.

\section*{Acknowledgments}
We thank all the anonymous reviews and Joakim Nivre who give a lot of valuable and insightful comments. 
We appreciate the grants provided by Erasmus+ Programme and Anna Maria Lundin's scholarship committee. 
GT is funded by the Chinese Scholarship Council (grant number \texttt{201607110016}). MM, AR and RS have received funding from the Swiss National Science Foundation (grant number \texttt{105212\_169888}).

\bibliographystyle{acl_natbib}
\bibliography{evaluation}

\end{document}